\documentclass[final,12pt,3p,times,twocolumn,authoryear]{elsarticle}

\usepackage{amssymb}
\usepackage{url}

\journal{Web Semantics}

\begin{document}

\begin{frontmatter}



\title{Comparison of biomedical relationship extraction methods and models for knowledge graph creation}

 \affiliation[nikola]{organization={University of Manchester, Faculty of Science and Engineering},
             addressline={Oxford road},
             city={Manchester},
             postcode={M13 9PL},
             country={United Kingdom}}
              \affiliation[nikola2]{organization={Bayer Pharmaceuticals R\&D},
             addressline={Mullerstrasse 178},
             city={Berlin},
             postcode={13353},
             state={},
             country={Germany}}   
 \author[nikola,nikola2]{Nikola Milosevic}

\author[wolfgang]{Wolfgang Thielemann}

\affiliation[wolfgang]{organization={Bayer Pharmaceuticals R\&D},
            addressline={Friedrich-Ebert-Str.475}, 
            city={Wuppertal},
            postcode={42117}, 
            state={},
            country={Germany}}

\begin{abstract}
Biomedical research is growing at such an exponential pace that scientists, researchers, and practitioners are no more able to cope with the amount of published literature in the domain. The knowledge presented in the literature needs to be systematized in such a way that claims and hypotheses can be easily found, accessed, and validated. Knowledge graphs can provide such a framework for semantic knowledge representation from literature. However, in order to build a knowledge graph, it is necessary to extract knowledge as relationships between biomedical entities and normalize both entities and relationship types. In this paper, we present and compare a few rule-based and machine learning-based (Naive Bayes, Random Forests as examples of traditional machine learning methods and DistilBERT, PubMedBERT, T5, and SciFive-based models as examples of modern deep learning transformers) methods for scalable relationship extraction from biomedical literature, and for the integration into the knowledge graphs. We examine how resilient are these various methods to unbalanced and fairly small datasets. Our experiments show that transformer-based models handle well both small (due to pre-training on a large dataset) and unbalanced datasets. The best performing model was the PubMedBERT-based model fine-tuned on balanced data, with a reported F1-score of 0.92. The distilBERT-based model followed with an F1-score of 0.89, performing faster and with lower resource requirements. BERT-based models performed better than T5-based generative models.

\end{abstract}


\end{frontmatter}


\section{Introduction}

The amount of published scientific, especially biomedical literature is growing exponentially. In 2020, over 950,000 articles were added to Medline \citep{Medline2020}, a repository of the biomedical literature, meaning that on average, over 2600 biomedical articles were published daily. Scientists, researchers, and professionals are not able to cope with the amount of published research in their area and require tools that would help them find relevant articles, review and validate claims and hypotheses.  

Finding relevant articles is the task addressed by the information retrieval sub-field of natural language processing. A number of information retrieval approaches have been examined and several domain-specific information retrieval applications for bio-medicine have been built, such as PubMed, PubMedCentral, Quertle, Embase, etc. \citep{canese2013pubmed,roberts2001pubmed,coppernoll2011quertle}. Information retrieval engines can use named entity recognition algorithms and entity normalization techniques (often using dictionaries or terminologies) to improve the search results by returning semantically the most relevant articles for the searched concept regardless of the form or a synonym used for the searched entity \citep{jonnalagadda2010nemo,jonnagaddala2015recognition,hakala2016syntactic}. However, information retrieval only offers a list of relevant articles for searched terms or concepts. In order to validate a hypothesis or claim, a researcher still needs to read through a significant amount of literature, which may be time-consuming. 

The hypothesis and claims, that researchers often would like to validate, can be summarized in a simple sentence with two interacting concepts and a predicate describing their interaction (e.g. \underline{Aspirin} \textit{treats} \underline{pain}). Hypothesis and claims are named relationships between concepts. These named relationships can be extracted with evidence (sentences from articles, stating them) from biomedical literature. Also, entities may be connected with many other entities in relationships, finally generating a large knowledge graph. This knowledge graph can be later utilized to infer knowledge by following connections ($A \rightarrow B, B \rightarrow C, therefore   A \rightarrow C$), and even applications of graph machine learning to find potentially missing edges (relationships), or discover potential leads and targets in the drug discovery process. 

The ways of validating claims, and inferring new knowledge from the statements in the knowledge graphs have been examined in areas of knowledge graph databases \citep{messina2017biograkn,miller2013graph}, semantic web \citep{mcguinness2004owl,parsia2004pellet,sirin2007pellet,shearer2008hermit} and graph machine learning \citep{scarselli2008graph,velivckovic2017graph,qu2019gmnn}. However, in order to perform inference and validation over a knowledge graph, the knowledge needs to be extracted from the text and normalized. Most of the normalization research in the biomedical domain considers the normalization of named entities, such as diseases, genes, and compounds \citep{cho2017method,ji2020bert,zhou2020knowledge}. On the other hand, it was not given much attention to the normalization of biomedical relationships. Relationship extraction research in the biomedical domain is often limited to a certain domain (e.g. cancer or cardiovascular domain) and considers a limited set of possible relationship entity pairs and relationship types \citep{rindflesch1999edgar,yang2021mining}. Normalized relationships (graph edges) are the pillar of successful systematization of knowledge in knowledge graphs.

As part of the R\&D organization within Bayer pharmaceuticals, we focus on generating knowledge graphs relevant to drug discovery, target identification, and indication expansion. Most of the use-cases we are dealing with are related to humans. Therefore, the defined relationship model and methods for relationship extraction, described in this paper, will focus on the stated use-cases.    

In this paper, we propose a data model for relationship normalization between drugs, targets, and diseases. We also examine and compare several rule-based and machine learning-based approaches. Using the proposed methods, we generated a knowledge graph with links to the evidence sentences, based on the extracted and normalized relationships from PubMed. In the end, we compare and discuss proposed methods for knowledge graph creation.

\section{Background}

Knowledge graphs have a long history spanning to the 1970s \citep{schneider1973course}. Knowledge graphs are a flexible knowledge representation framework, where knowledge is represented as a graph of inter-related concepts. Representing knowledge in a graph has a number of practical benefits in scenarios that involve integrating, managing, and extracting value from diverse and heterogeneous data sources. The idea of representing knowledge in graphs, particularly gained influence with Semantic Web, and lately with the development of knowledge graph announced in 2012 by Google, followed by other major tech industry players \citep{hogan2021knowledge}. Lately, we could see applications of knowledge graphs in question answering products in wider use, such as Alexa, Google Assistant, or Siri \citep{zhang2018variational}. Likewise, the pharmaceutical industry identified potential benefits knowledge graphs can bring in accelerating drug discovery, drug development, indication expansion of existing drugs, and pharmacovigilance.

In order to extract information and structure them for entry into the knowledge graph, it is necessary to perform named entity recognition of relevant entities (for bio-medicine these could be genes/targets, compounds, diseases, cell lines, pathways, organs, etc.), normalize all the possible synonyms to agreed terminology and at the end find relationships between co-mentioned entities and normalize the relationships to the agreed data model or ontology.

Biomedical named entity recognition and named entity normalization have a long tradition of research since the late 1990s \citep{fukuda1998toward,collier2000extracting}. A number of approaches were developed that can be classified into dictionary-based, machine learning-based using usually hidden Markov models or Conditional Random Fields and Deep Learning-based, often using language models, such as word2vec, ELMo \citep{milosevic2020mask}, BERT, and others, with transformers \citep{khan2020mt} or recurrent neural networks \citep{belousov2019extracting}. 

In order to systematize extracted entities and input them into the knowledge graph, they need to be normalized. Normalization is a process of mapping all possible terms and variants that represent one concept to one unique entity id or preferred term (for example the concept of cancer can be stated using various expressions, such as neoplasms, tumor, cancer, malignity, etc.).  For a long time, named entity normalization relied on good dictionaries and rule-based approaches \citep{leaman2015tmchem,cohen2005unsupervised}. However, in recent years, there have been several deep learning-based approaches for ranking candidate entities for normalization using convolutional neural networks \citep{li2017cnn,deng2019ensemble} or language models such as Word2Vec \citep{cho2017method} or BERT \citep{ji2020bert}. 

The extraction of the actual relationship comes as the final step of structuring information from a sentence. One of the first systems to attempt relationship extraction in the biomedical domain was EDGAR \citep{rindflesch1999edgar}, which was extracting relationships between drugs and genes in the cancer domain, using a set of rules based on syntactic analysis. Since then, several approaches to structuring relationships were explored: 
\begin{itemize}
    \item \textbf{Existence of relationship between entities} - classifies whether there is an actual semantic relationship between two entities, or the entities are co-mentioned, but there is no actual named relationship between them. This approach of extracting general existence of the relationship is often applied for protein-protein interaction extraction \citep{zitnik2018modeling,szklarczyk2019string} or gene-disease interactions \citep{becker2004genetic}.
    \item \textbf{Extracting predicate verb as relationship type} - predicate is not a closed set of possible classes. Any predicate verb, appearing in a sentence indicating a relationship between entities, is taken as the relationship type. Normalization of relationship types is left for further processing or analysis. Some of the research databases, such as Open Targets use this approach \citep{carvalho2019open}, as well as some of the commercial tools that allow relationship extraction (e.g. Linguamatics I2E). 
    \item \textbf{Normalizing relationship types} - predicate is normalized into the set of well-defined types. This approach needs a carefully crafted data model of possible relationships, as well as a carefully developed dataset for machine learning or extraction rules. In the semantic web community, there has been research on normalizing a basic set of relationships in the general domain, such as is-a, part-of, equal \citep{arnold2015semrep,speer2013conceptnet,speer2017conceptnet}. Domain-specific and more granular data models and datasets for this approach are rare. BioCreative VI and BioCreative VII provided data and organized shared tasks on chemical-protein interactions \citep{krallinger2017overview,krallingerdrugprot}
\end{itemize}

From the methodological perspective, relationship extraction can be performed using machine learning or rule-based approaches. Rule-based approaches range from using lists of relationship-related keywords and distances between concepts and keywords \citep{abacha2011automatic,ravikumar2017belminer} to using dependency parsers and evaluating whether concepts are related in grammatical sense \citep{erkan2007semi,goertzel2006using}. On the other hand, machine learning approaches can be classified into two groups: (1) supervised learning, using crafted datasets \citep{peng2018chemical,liu2017attention} and (2) semi-supervised or distant learning approaches, where a dataset is expanded based on known relationships assuming that mentions of the same entities would entail the same relationship \citep{mintz2009distant}. Since 2009, distant learning approaches have gained popularity and proved to be effective in relationship extraction, despite the assumption that all co-mentions of the same entities would entail the same relationship is not always correct and adds noise. In addition to these two general approaches, hybrid approaches, combining rules, dependency trees, and machine learning approaches have been popular for relationship extraction \citep{erkan2007semi,muzaffar2015relation}.

Over the last decade, several datasets for supervised biomedical relationship extraction have been developed.  Some of the widely adapted and used datasets for relationship extraction are BioCreative VI and VII Chem-Prot datasets \citep{miranda2021overview}, BioCreative V Chemical-disease (CDR) dataset \citep{li2016biocreative}, ADE Corpus \citep{gurulingappa2012development}, BioInfer \citep{pyysalo2007bioinfer}, DDI'13 \citep{herrero2013ddi}, n2c2 2018 ADE \citep{henry20202018}, N-ary \citep{peng2017cross}, BioRED \citep{luo2022biored} and others. However, few of these datasets have granular manually curated relationship types annotated (many have either binary marker, or broad relationships, or are not in the scope of this paper - e.g. drug-drug or protein-protein interactions are out of the scope of this study).

\section{Method}

In this paper, we compare methods for relationship extraction. All of our methods use in-house modified Linnaeus  \citep{gerner2010linnaeus} tool for named entity recognition and normalization. For relationship extraction, we present and compare three methods: (1) a rule-based method, based on sentence patterns and dictionaries of trigger verbs and phrases,  (2) a machine learning method, based on traditional machine learning models (i.e. Naive Bayes, Random Forests), and (3) a deep learning method based on transformer architectures, such as DistilBERT \citep{sanh2019distilbert}, text to text T5 transformer \citep{raffel2020exploring}, as well as domain-specific BERT-based model called PubMedBERT \citep{gu2021domain} and domain-specific version of T5 model, called SciFive \citep{phan2021scifive}.

\subsection{Named entity recognition and normalization (modification of Linnaeus)}

Named entity recognition was done by an internally modified version of the Linnaeus tool \citep{gerner2010linnaeus}. We have added a number of features that would allow us to perform more flexible entity matching while relying on the Linnaeus algorithm which is fast and reliable with good dictionaries. The added features include: 
\begin{itemize}
    \item Handling a defined set of characters, such as white spaces and treating multiple sequential white space characters as a single white space character. 
    \item Implementing a global flag that allows ignoring cases of letters in matches if needed
    \item A flag for automatic pluralization (adding \textit{-s} and \textit{-es} suffixes) of dictionary terms
    \item A functionality that can handle and transliterate Greek characters (e.g. beta - $\beta$) as well as functionality that can handle the variation of the position of Greek characters (e.g. Interferon-$\alpha$ vs $\alpha$ -interferon).
    \item Removing or ignoring diacritics
    \item Synonym level exact, case sensitive, and regular expression matching
    \item Flag to match only the longest match
\end{itemize}

The dictionaries used for named entity recognition and normalization into entities have been carefully internally developed, expanded, and refined over the past fifteen years by our internal information scientists. We have used dictionaries for human genes, diseases, and approved drugs.

\subsection{Relationship data model}

Relationships of interest are relationships between drug, gene, and disease entities. The three pairs of relationships of interest are: (1) Drug-Gene, (2) Drug-Disease, and (3) Gene-Disease. Each of these pairs may have several distinct relationship types. In order to develop the relationship model, we have organized two workshops guided by the authors with the internal experts from the Bayer R\&D department. Eighteen people participated in these workshops and the effort to create the data model. They are members of the following teams within the Bayer R\&D department: scientific and competitive intelligence (12 people), semantics and knowledge graph technologies (3 people, including authors), research and early development, kidney disease (2 people), bioinformatics (1 person). Experts from these teams have advanced degrees (Ph.D.) in pharmacology, biology, or medicine and often substantial working experience in academia and within the pharmaceutical industry. They have helped us identify the possible relationship types for the entities we focused on and validate our model. The discussions were guided by authors, who proposed the initial data model, and then it was evaluated, critiqued, and expanded by the group of experts. The authors were starting with existing data models if they existed (e.g. for gene-disease, BioCreative data model), or known requirements coming from the drug discovery department. These models were then simplified, reviewed and, in some cases, missing relationship types were added. We organized a meeting with one bioinformatics expert, who helped us additionally expand and validate Gene-Disease relationships and possible modes of action. Furthermore, we scouted available commercial solutions that provide knowledge graph solutions with entities we were looking for. We have identified two companies which provide data that is close to our needs - Causally\footnote{\url{https://www.causaly.com/}} and Biorelate\footnote{\url{https://www.biorelate.com/}}. The model, that we created, contained more comprehensive and more detailed relationship types (more relationship types, relationship attributes, such as modes of action for genetic relationships) for the relevant entity pairs, at the time of writing this paper.

\subsubsection{Drug-Gene relationships}
\begin{table}[!htbp]
    \centering
    \begin{tabular}{|l|l|}
    \hline
        CPR GROUP & TYPE \\ \hline
        CPR:1 & Part of \\
        CPR:2 & Regulator\\
        CPR:3 & Up-regulator, Activator\\
        CPR:4 & Down-regulator, Inhibitor\\
        CPR:5 & Agonist\\
        CPR:6 & Antagonist\\
        CPR:7 & Modulator\\
        CPR:8 & Co-factor\\
        CPR:9 & Substrate, Product of\\
        CPR:10 & Not\\
        \hline
    \end{tabular}
    \caption{Chemical-Protein relationship as defined by BioCreative shared task}
    \label{tab:cpr_groups}
\end{table}

The relationships between chemical and proteins were subject of last two BioCreative shared tasks in 2017 \footnote{\url{https://biocreative.bioinformatics.udel.edu/tasks/biocreative-vi/track-5/}} and 2020 \footnote{\url{https://biocreative.bioinformatics.udel.edu/tasks/biocreative-vii/track-1/}}. Both of the tasks defined the same interaction types. These can be seen in Table \ref{tab:cpr_groups}. However, for the majority of purposes, some of the defined types may be redundant. Therefore, we have simplified the data model by merging some of the classes and excluding ones that are rarely mentioned in the text. In the end, our Drug-Gene model had the following relationship classes: 
\begin{itemize}
    \item Up-regulator/activator
    \item Down-regulator/inhibitor
    \item Regulator
    \item Part of
    \item Modulator
    \item Co-factor
    \item Substrate or product of
\end{itemize}

Note that \textit{Regulator} is a type of relationship in which it is not possible to determine the direction of regulation from the sentence.

\subsubsection{Drug-Disease relationships}

Relationships between drugs and diseases do not have any gold standard data model that was previously used. Therefore a new model was proposed containing the following relationship classes: 
\begin{itemize}
	\item Therapeutic use/Treatment
	\item Cause/Adverse event
	\item Decrease Disease
	\item Increase Disease
	\item Effect on
	\item Biomarker
\end{itemize}

It may seem that there is redundancy between \textit{Therapeutic use} and \textit{Decrease Disease} classes, or between \textit{Cause} and \textit{Increase Disease} classes. However, \textit{Therapeutic use} and \textit{Cause} indicate relationships where the disease is an indication or counter indication for a given drug. Increase and Decrease disease may refer to any finding that a given drug improved or worsened the state of disease, and therefore is a weaker relationship. The weakest relationship is \textit{Effect on}, because, in this case, only the fact that there is some effect of a drug on disease is known, without any additional details (e.g. whether it improves disease or makes it worse).  

The chemical compound can be a biomarker for some diseases. In medicine and drug discovery, it is important to have a picture of biomarkers, and therefore it is included in the model.

\subsubsection{Gene-Disease relationships}

The relationship between genes and diseases is the most complex one among the three types in scope. This is because a single gene can improve, worsen or even cause a certain disease. Therefore, it is often not enough to classify the type of the relationship, the algorithm needs to extract also a mode of action on the gene. In terms of possible relationship types, we have identified the following ones: 
\begin{itemize}
	\item Plays a role –- From the sentence can be concluded clearly that there is a connection between the gene and the disease, however, it is not clear what kind of role the gene plays in the disease, only that it plays some role. 
	\item Target
	\begin{itemize}
		\item General –- The gene or protein can be considered a target for the given disease, with no more specific information on the modulation of the disease.
		\item Cause –- The sentence indicates that activation, mutation or inhibition, or any other action over a gene is causing a given disease.
	\item Modulator
	\begin{itemize}
	\item Decrease disease -– There is a clear indication that gene is responsible for decreasing and alleviating the disease. 
	\item Increase Disease -- There is a clear indication that gene is responsible for increasing and worsening the disease.
	\end{itemize}	
	\end{itemize}

	\item 	Biomarker –- The presence or lack of a given gene/protein is an indicator for the diagnosis of disease or pathology.

\end{itemize}

\subsubsection{Mode of action}

Together with the relationship classes, if available, mode of action is an important modifier for Gene-Disease relationships. It establishes the action taken on a gene in order for the relationship to take place. For example, a gene may both decrease and cause disease, depending on whether the gene was activated or inhibited. Possible modes of action are (1) inhibition or down-regulation, (2) activation or up-regulation, (3) mutation or modification.

\subsubsection{Negation and speculation}

The relationship between entities in a sentence may be negated, which reverses the semantics of the relationship. Therefore, it is important to detect whether the relationship is negated.

Likewise, statements in the text can be factual, stated as well-known facts, or speculative. Speculative claims need to be taken with more caution and therefore, speculation detection is included in our model.

\subsection{Relationship extraction using rule-based method}

\begin{figure*}[h]
\caption{Example of dictionaries, rules set and an example of sentence annotations in order to match relationship in a sentence}
\centering
\includegraphics[width=1\textwidth]{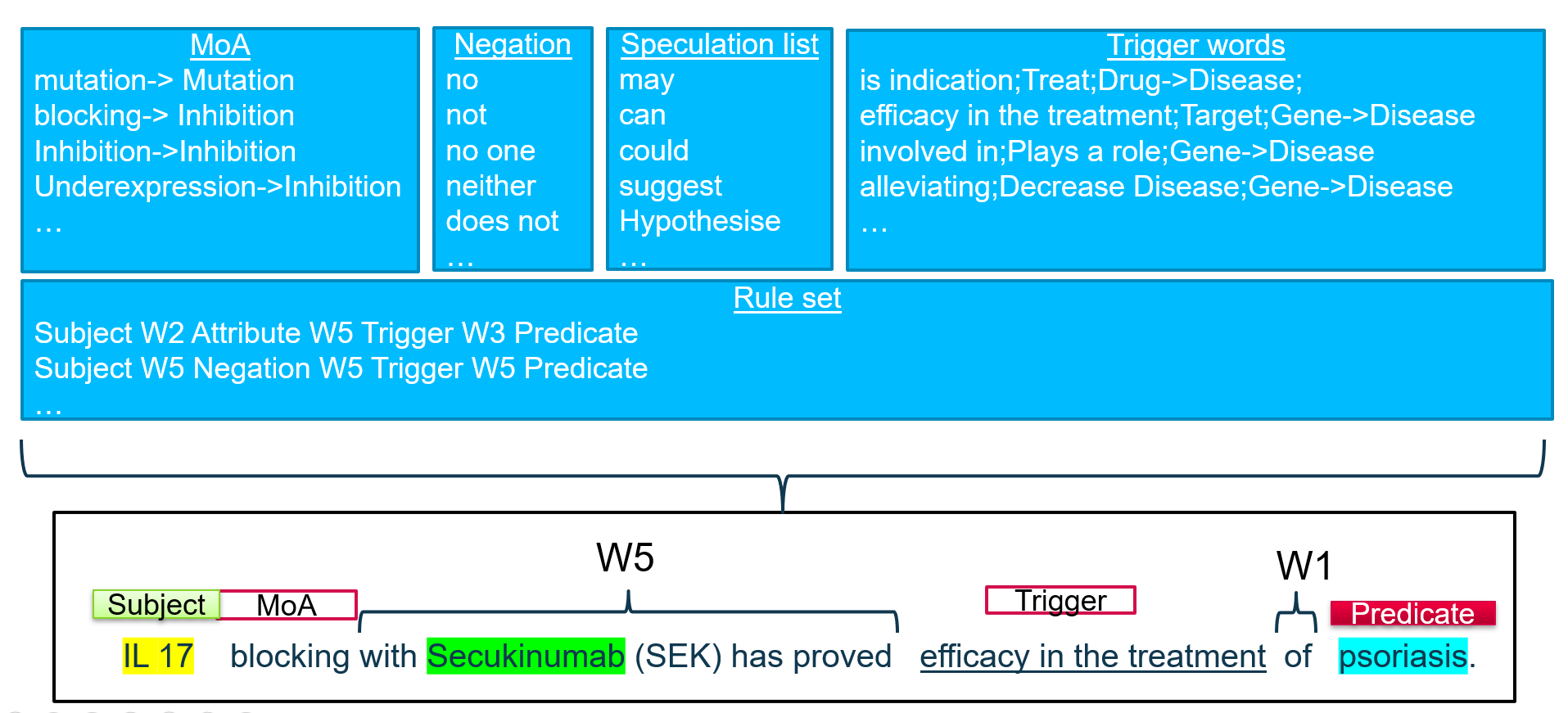}
\label{fig:RuleBased}
\end{figure*}

Based on the previously described relationship model, we have developed a rule-based method for relationship extraction. The method relies on vocabularies for relationship trigger words, negation cues, speculation cues, mode of action (MoA) cues, and grammar pattern rule set. An example of vocabularies and patterns in the ruleset with an example sentence from which a relationship is extracted using given rules and vocabularies is presented in Figure \ref{fig:RuleBased}.

The trigger word vocabulary contains a list of relationship trigger words and phrases, with metadata, such as to which relationship a given word or phrase maps, between which entities, whether entities have to be in a given order of mentioning or can be in reverse order (e.g. for Drug-Disease relationship whether it is allowed for the drug to be after disease in the sentence) and whether the phrase should be interpreted as a regular expression.

Mode of action cues has a mapping to the mode of action type (i.e. Inhibition, Activation, Mutation). The vocabularies for negation and speculation cues are simple lists of words (e.g. \textit{no, not} for negation; \textit{hypothesise, may} for speculative). 

Grammar patterns define sequences that need to be matched in order to extract relationships. This grammar has several keywords, such as \textit{Subject}, which refers to the subject entity, \textit{Predicate}, which refers to predicate entity, \textit{Trigger}, referring to trigger cue, \textit{Speculation}, referring to Speculation, \textit{Negation}, referring to negation cue, \textit{Subject\_type}, referring to entity that is not subject in current evaluation pair of entities, but has same type as Subject entity (i.e. Drug or Gene), \textit{Predicate\_type}, referring to entity having same type as predicate, but not evaluated in current pair. Additionally, there may be defined a number of words that are between labeled entities, trigger words, negations, and speculative phrases. For example the following pattern:

\begin{quote}
\textit{Speculation W3 Subject W3 Trigger W3 Predicate}
\end{quote}

would match sequences where the speculative cue is followed by up to three tokens, followed by Subject, followed by up to three tokens, followed by trigger phrase, followed by another three tokens and predicate. This means it would match sentences such as "We hypothesize that aspirin can alleviate most cases of headache", if the token "hypothesize" is in the list of speculative cues, "aspirin" is marked as a drug and is subject, "headache" is a disease and predicate, and "alleviate" is a trigger word. 

The matching algorithm iterates over labeled entities in each sentence and finds all pairs that may constitute relationships. It annotates sentences with potential trigger phrases, speculative cues, and negations. Finally, the algorithm tries to match any pattern from the grammar to the sequence in a sentence. If the matching is successful, the relationship is extracted and mapped to the relationship type and metadata, such as mode of action, negation, and speculation cues are extracted. 

The confidence score is calculated as a proportion of words in sentences that exactly match named phrases (Subject, Predicate, Trigger, Speculation, Negation), divided by all the words in the pattern (this includes named phrases and tokens that were matched as part of allowed distance tokens, e.g. up to three words for each W3 statement in grammar). The rationale for this calculation is based on the assumption that additional words may change the semantics of the sentence and therefore confidence about the existence of the relationship should be lower. 

In addition, the method also extracts co-occurrences, giving them fixed confidence of 0.0001, and labeling them with a "Co-occurrence" label. 

Each extracted relationship contains information about entities (entity string, type, preferred term, internal ID), relationship type, whether it is negated, whether it is speculative, and confidence score. Also, the evidence sentence and Medline ID of the article where the evidence was found are recorded.

\begin{table*}[h!]
    \centering
    \begin{tabular}{|l|r|r|}
    \hline
        Relationship type & Unbalanced dataset & Balanced dataset  \\ \hline
         Biomarker & 198 & 1243  \\
       No Explicit Relationship   & 446 & 446 \\
         Plays a role & 7393 & 1532 \\
         Target-\textgreater Causative  & 1460 & 1508 \\
        Target-\textgreater General        & 656 & 1414  \\
         Target-\textgreater Modulator-\textgreater Decrease Disease & 1108 & 1450  \\
         Target-\textgreater Modulator-\textgreater Increase Disease   & 720 & 1422  \\
        \hline
        
    \end{tabular}
    \caption{Number of sentences for each relationship type in our balanced and unbalanced data sets}
    \label{tab:dataset_stats}
\end{table*}

\subsection{Machine learning}

We have developed machine learning methods for classifying relationship types. The task was modeled as a sentence-level classification task.  The initial method is based on sentence classification using traditional machine learning algorithms, such as Random Forests, and Naive Bayes. We have then advanced the method by using fine-tuned transformer-based architectures for sentence-level relationship type classification, such as DistilBERT \citep{sanh2019distilbert}, and a text-to-text transformer called T5 \citep{raffel2020exploring}. We have also compared these methods to domain-specific variants of BERT and T5 models, called PubMedBERT \citep{gu2021domain} and SciFive \citep{phan2021scifive} respectively.  We report here results from all of the mentioned experiments.

\subsubsection{Training and testing data}

The data are collected by using a rule-based relationship extractor previously described for the Gene-Disease relationship. We are evaluating our approach to Gene-Disease data, as it proved to have the most complex data model and therefore is the most complex to correctly extract relationships. Also, this relationship type is important from a biomedical perspective, as it may give insights on potential targets for treating respective diseases. From this dataset, 2000 sentences were reviewed and corrected by human annotators. For this task, a company called Molecular Connections was contracted. Other 10 000 sentences were obtained from the rule-based method, with confidence 1. These sentences would match correct sentences, as they do not allow for any tokens that may change context, apart from named phrases. Therefore, the dataset contained about 12 000 sentences. The data was split as 90\% training and 10\% testing data for training and testing of machine learning approaches.

In order to create a more balanced dataset, we have generated the second dataset by taking 2000 manually annotated sentences, but then adding sentences from the rule-based method with confidence 1 in such a way that each relationship class had at least 1400 sentences (for biomarkers, we could obtain 1243 sentences with confidence 1, from a processed portion of the data we had at the time of building the dataset). The statistics about the number of sentences per relationship class in our datasets are presented in Table \ref{tab:dataset_stats}.

We have also created a dataset for the classification of a mode of action. We created again one unbalanced and balanced dataset. Since for the mutation class we had only 140 examples, we initially balanced the dataset by taking 140 examples from each class. This is a fairly small dataset and may be improved by adding examples. We have created an additional dataset taking 300 data samples from each class, allowing duplication for classes that did not have enough samples (e.g. mutation). Since \cite{tarawneh2022stop} argued that oversampling with fictitious data may result in the model failing when put to real-world problems, we perform only duplication, keeping just real-world data.  Access to the generated datasets can be requested at \url{https://zenodo.org/record/6466316#.Ylw3T-dS9Ea}.

\subsubsection{Initial experiments using Random Forest and Naive Bayes classifiers}

Initially, we attempted to use traditional machine learning algorithms, such as Naive Bayes and Random Forests. For both of them, sentences were tokenized and stemmed using Porter Stemmer \citep{porter1980algorithm}. Since for relationship extraction, it is important to examine the sequences of words, the features for our classifiers were uni-grams, bi-grams, tri-grams, and four-grams. Finally, data was trained using Random Forest and Naive Bayes Classifier.

\subsubsection{Transformer-based architectures: DistilBERT, PubMedBERT and T5 and SciFive-based models}

Transformer-based models are currently the state-of-the-art machine learning methods that perform well on a variety of tasks, ranging from classification to summarization and question answering \citep{devlin2018bert,raffel2020exploring}. In the past few years, a number of language models were developed and pre-trained on datasets such as common crawl. Many of these models are based on the BERT model, with various modifications to reduce the size of the model or increase speed \citep{sanh2019distilbert,liu2019roberta,lan2019albert}. These models can be used for classification by using and training head - a feed-forward neural network on top of the transformer network that outputs predictions. We will use a BERT-based model that was optimized for size and speed, called DistilBERT \citep{sanh2019distilbert}, whose authors claimed that has 40\% fewer parameters, runs 60\% faster while preserving over 97\% of BERT's performances as measured on the GLUE language understanding benchmark. Both BERT and DistilBERT are trained in the general domain. Since our task is specific to the biomedical domain, we will also use a BERT-based model trained on PubMed and PubMed Central (PMC) articles released by Microsoft, called PubMedBERT \citep{gu2021domain}. In 2020, Google released a text-to-text transformer called T5. This model is generating textual output and a single model can be trained to perform multiple tasks (specifying task in the prefix of the input). In the original paper, the authors of T5 claimed that the model exhibits state-of-the-art performance and on most of the tasks it outperformed BERT. In this paper, we will evaluate that claim on the sentence-level classification of biomedical relationships (gene-disease). We will also compare it to the biomedical variant of the model called SciFive \citep{phan2021scifive}.

The learning task was defined as a sentence classification task. For a given sentence, containing entities, the model is supposed to provide a normalized relationship type from our data model. In the training and testing sentences, the text was pre-prepared in such a way that subject of the relationship (e.g. gene) was masked with the keyword \textbf{SUBJECT} and the predicated of the relationship (e.g. disease) was masked with the keyword \textbf{PREDICATE}. In this way, we hypothesized that the internal attention mechanism of the model would be able to learn how to treat the vicinity of subjects and the predicates of the relationships. 

The DistilBERT model was based on the DistilBERT base uncased model available on HuggingFace\footnote{https://huggingface.co/distilbert-base-uncased}. This model was fine-tuned for the classification task, and trained on our unbalanced and balanced datasets for 5 epochs (learning rate=0.00002). DistilBERT is an encoder model, to which a decoder can be created using a pooling and feed-forward network whose output layers dimension is equal to the number of classes (in our case 8).

Similarly, PubMedBERT was based on the base uncased version of the model trained on both PubMed and PMC data that is available on HuggingFace\footnote{https://huggingface.co/microsoft/BiomedNLP-PubMedBERT-base-uncased-abstract-fulltext}. The model was also trained for 5 epochs and the same configuration was used for learning rate, batch size, and sequence size as for DistilBERT model.

On the other hand, the T5 model has encoder-decoder architecture, and therefore we do not define additional layers. We have fine-tuned the T5 model that is readily available on HuggingFace\footnote{https://huggingface.co/t5-base}. T5 is a multi-task model that can be fine-tuned and new tasks can be added during the fine-tuning of the model. The multi-tasking nature of T5 is convenient since the same model can be deployed once performing multiple tasks (e.g. question-answering, summarization, translation, and relationship extraction within the same API). During the fine-tuning of the model, we have added new prefixes for gene-disease relationship classification and gene mode-of-action classification (with four classes - activation, inhibition, mutation, and not reported). We have fine-tuned the model on our dataset using Adafactor optimizer \citep{shazeer2018adafactor}. The model was trained for 5 epochs (learning rate=0.00002). Same fine-tuning was performed on the domain-specific SciFive base model available on HuggingFace\footnote{https://huggingface.co/razent/SciFive-base-Pubmed\_PMC}. 

Encoder layers of T5, SciFive, and PubMedBERT have a size of 512 tokens, while DistilBERT has an encoder size of 728 tokens. Sequence sizes are longer than the longest sentence in our dataset, therefore the size difference should not affect the training and we used padding to fill the sequence with special padding tokens.

\section{Results}

\subsection{Rule-based relationship extraction}

We have processed base Medline data until January 2021, containing about 35 million abstracts. The processing with both Linnaeus and the relationship extraction engine took about 7 days on a single machine. We managed to extract in total 4,784,985 relationships (with co-occurrences 35,900,521). There were 631,573 named relationships found between Drug-Genes (6,885,810 including co-occurrences), 1,468,639 relationships between Drug-Diseases  (8,378,599 including co-occurrences), and  2,684,742 relationships between Genes and Diseases (20,065,385 including co-occurrences). 

  \begin{figure*}[h]
\caption{Section of knowledge graph showing nodes that are in relationship with autosomal dominant polycystic kidney disease (ADPKD). Orange entities are diseases (ADPKD), entities in blue are drugs and in green are genes/proteins.  Label on edges present relationship type, number of mentions and cumulative confidence score for the given relationship between two entities.  }
\centering
\includegraphics[width=1\textwidth]{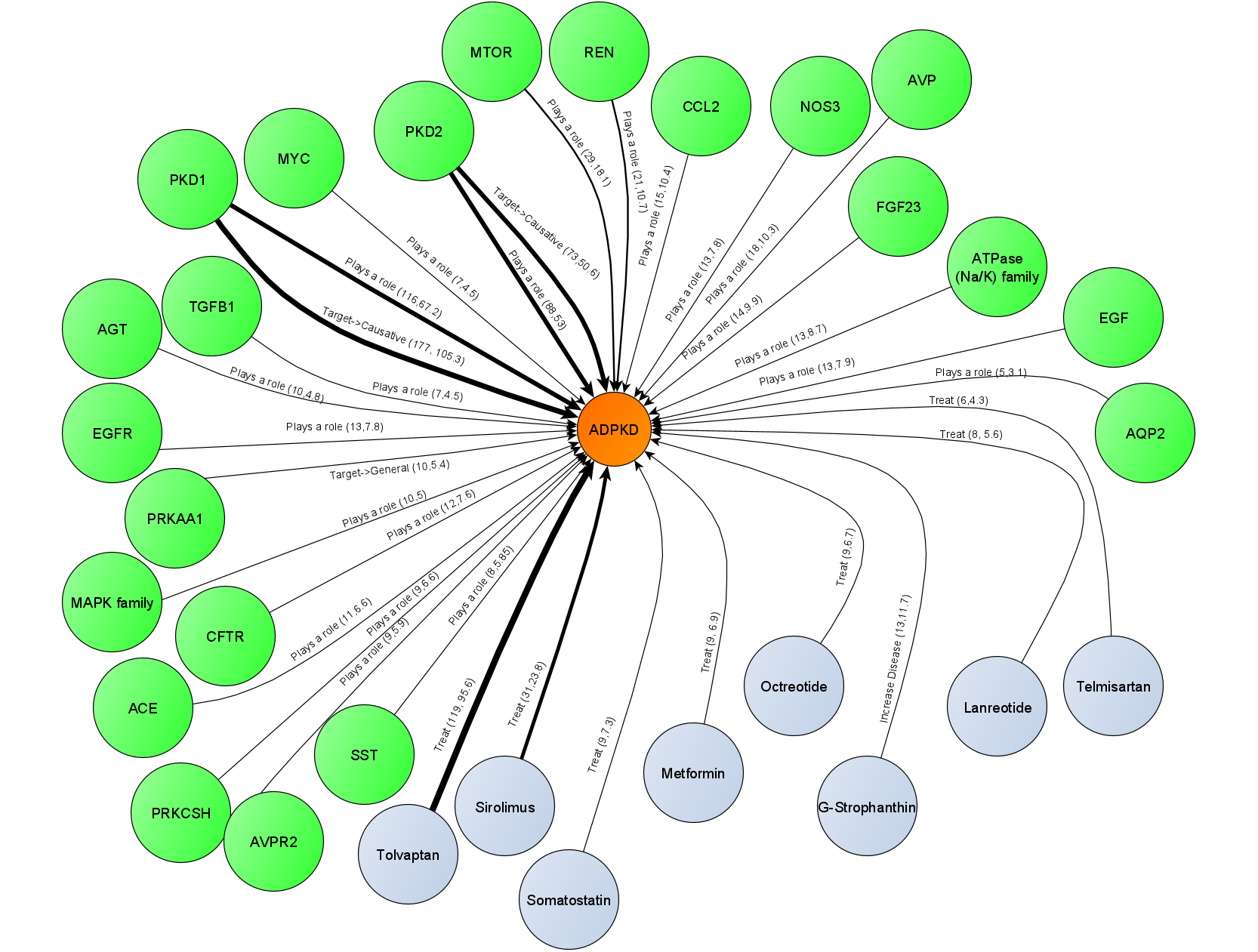}
\label{fig:ADPKDKG}
\end{figure*}

Extracted relationships can be loaded into a graph or relational database, where these relationships can answer complex medical questions with evidence. By summing confidence scores, it is possible to retrieve genes interacting with a given drug (e.g. top results for drug \textit{Tolvaptan} was inhibition of AVPR2, while the second one was inhibition of vasopressin receptor family), drugs that have an effect on certain disease (e.g. for \textit{autosomal dominant polycystic kidney disease} retrieved Tolvaptan, which is approved for autosomal dominant polycystic kidney disease, Sirolimus, which inhibits mTOR and as well have been often used in polycystic kidney disease, and Somatostatin, which was published as a hormone having a potential role in the treatment of autosomal dominant polycystic kidney disease \citep{messchendorp2020somatostatin}), or what genes are important for a given disease (e.g. for autosomal dominant polycystic kidney disease retrieved PKD1 and PKD2 as targets that both play a role and have a causative relationship with disease, as well as mTOR, REN, CCL2). We evaluated a case study related to autosomal dominant polycystic kidney disease. We created a graph whose edges end in autosomal dominant polycystic kidney disease. In order to reduce noise, we consider only edges that represent the relationship that was mentioned at least 5 times in PubMed. We then evaluated the graph and all entities were indeed known from the literature to experts in the kidney disease team. A portion of the knowledge graph with relationships ending in ADPKD is presented in Figure \ref{fig:ADPKDKG}.

We have manually evaluated 100 abstracts containing at least one relationship and calculated precision, recall, and F1-score. The evaluation is depending on the extent of trigger phrases and completeness of grammar, which is overtime improving. The measured performance was 0,88 precision, 0,74 recall and 0,80 F1-score. It is expected that the rule-based approach would have high precision and lower recall, as it would miss some of the relationships, but annotate relatively few false positives. Despite making some false positive relationships, generated data perform well in answering relevant biomedical questions between genes, drugs, and disease. The cumulative effect is that noise can be ignored by setting a threshold and manually validating results below the given threshold if necessary.

\subsection{Naive Bayes and Random Forests-based Relationship extraction}

  \begin{figure*}[h]
\caption{F1-score by epoch in fine-tuned DistilBERT and T5 models on both unbalanced and balanced datasets }
\centering
\includegraphics[width=1\textwidth]{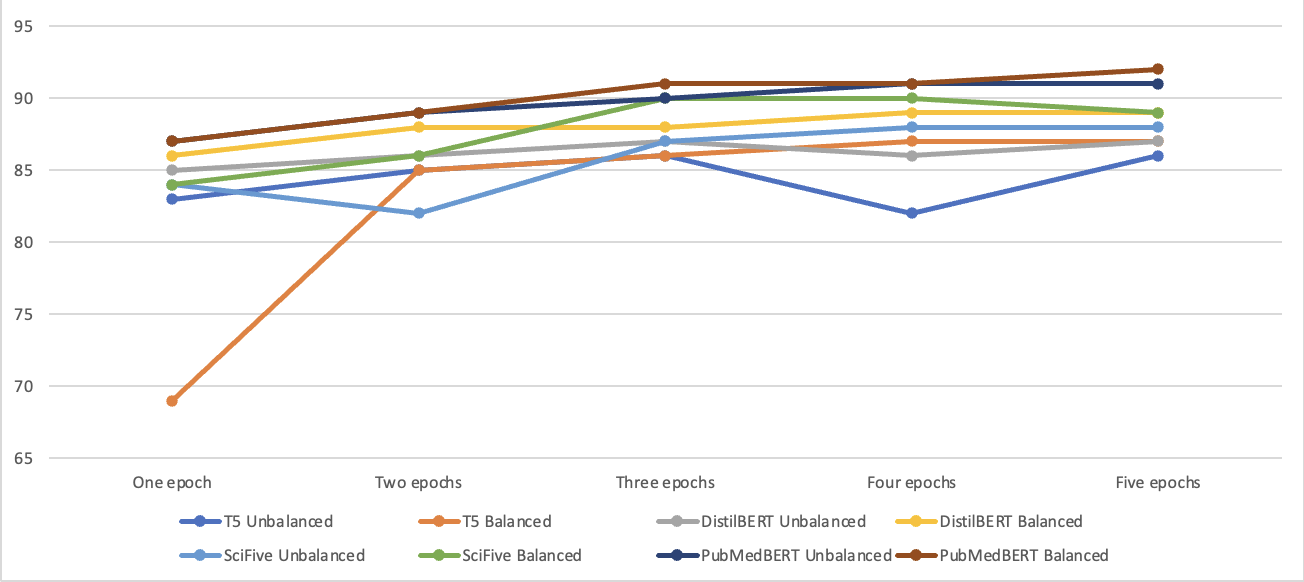}
\label{fig:t5_epochs}
\end{figure*}
The machine learning method was evaluated on two sets (2000 manually annotated relationships + 10,000 random relationships with confidence 1 - unbalanced set, and 2000 manually annotated relationships + random relationships with confidence 1, so there are at least 1,500 examples for each class - balanced set). For both data sets, 90\% of data was used as training data, while 10\% of data (about 1200 sentences) was used as a testing set. The results of our evaluation can be seen in the table \ref{tab:trad_res}.

Balancing data significantly improves precision and recall in both classifiers. With unbalanced data, Naive Bayes learned to always pick the most probable class - the class with the most results. The random forest classifier was better at learning how to recognize classes. However, balancing data, gained 26\% in F1-score for Naive Bayes and 14\% for overall results in Random Forests. The worst performance has a class which we were unable to balance due to the lack of annotated examples - \textit{No Explicit relationships} (477 sentences in the unbalanced set, that was annotated by annotators). Other classes performed with an F1-score over 70\%. 

\begin{table*}[!]
    \centering
    \begin{tabular}{|l|l|l|l|}
    \hline
        Class & Precision & Recall & F1-score \\ \hline
        \multicolumn{4}{|l|}{Unbalanced dataset}   \\  \hline
        \hspace*{3mm} Naive Bayes &  0.39 & 0.62 & 0.48 \\
        \hspace*{6mm} Biomarker & 0 & 0 & 0 \\
        \hspace*{6mm} No Explicit Relationship   & 0 & 0 & 0 \\
        \hspace*{6mm} Plays a role & 0.62 & 1 & 0.77 \\
        \hspace*{6mm} Target-\textgreater Causative  & 0 & 0 & 0\\
        \hspace*{6mm} Target-\textgreater General        & 0 & 0 & 0 \\
        \hspace*{6mm} Target-\textgreater Modulator-\textgreater Decrease Disease & 0 & 0 & 0 \\
        \hspace*{6mm} Target-\textgreater Modulator-\textgreater Increase Disease   &0 & 0 & 0 \\
        \hline
        \hspace*{3mm} Random Forests &  0.74 & 0.71 & 0.66  \\
        \hspace*{6mm} Biomarker & 0.80 & 0.16 & 0.27 \\
        \hspace*{6mm} No Explicit Relationship   & 0.89 & 0.30 & 0.45  \\
        \hspace*{6mm} Plays a role & 0.70 & 0.99 & 0.82 \\
        \hspace*{6mm} Target-\textgreater Causative  & 0.81 & 0.34 & 0.48 \\
        \hspace*{6mm} Target-\textgreater General        & 0.75 & 0.24 & 0.37\\
        \hspace*{6mm} Target-\textgreater Modulator-\textgreater Decrease Disease & 0.76 & 0.31 & 044 \\ 
        \hspace*{6mm} Target-\textgreater Modulator-\textgreater Increase Disease   &0.81  & 0.17 & 0.28 \\
        \hline
         \multicolumn{4}{|l|}{Balanced dataset}  \\  \hline
        
        \hspace*{3mm} Naive Bayes &  0.73 & 0.75 & 0.74 \\
        \hspace*{6mm} Biomarker & 0.94 & 0.91 & 0.92 \\
        \hspace*{6mm} No Explicit Relationship   & 0 & 0 & 0 \\
        \hspace*{6mm} Plays a role & 0.66 & 0.75 & 0.70 \\ 
        \hspace*{6mm} Target-\textgreater Causative  & 0.66 & 0.89 & 0.76\\
        \hspace*{6mm} Target-\textgreater General        & 0.83 & 0.73 & 0.78 \\
        \hspace*{6mm} Target-\textgreater Modulator-\textgreater Decrease Disease & 0.74 & 0.72 & 0.73 \\
        \hspace*{6mm} Target-\textgreater Modulator-\textgreater Increase Disease   &0.84 & 0.76 & 0.80 \\ 
        \hline
        \hspace*{3mm} Random Forests &  0.79 & 0.79 & 0.78  \\
        \hspace*{6mm} Biomarker & 0.97 & 0.85 & 0.91 \\  
        \hspace*{6mm} No Explicit Relationship   & 0.64 & 0.15 & 0.24  \\
        \hspace*{6mm} Plays a role & 0.65 & 0.80 & 0.72 \\
        \hspace*{6mm} Target-\textgreater Causative  & 0.81 & 0.87 & 0.84 \\
        \hspace*{6mm} Target-\textgreater General        & 0.76 & 0.84 & 0.80\\ 
        \hspace*{6mm} Target-\textgreater Modulator-\textgreater Decrease Disease & 0.75 & 0.81 & 0.78 \\  
        \hspace*{6mm} Target-\textgreater Modulator-\textgreater Increase Disease   &0.91  & 0.81 & 0.86 \\
        \hline
    \end{tabular}
    \caption{Results of Naive Bayes and Random Forests classifiers}
    \label{tab:trad_res}
\end{table*}

\subsection{Transformer-based relationship extraction}

We have fine-tuned base T5 and SciFive models for relationship extraction by adding a new prefix (\textit{"Relationship extraction:"}) on both unbalanced and balanced data. We have monitored the performance of the algorithm over epochs. The results can be seen in Figure \ref{fig:t5_epochs}.

Likewise, we have trained the DistilBERT and PubMedBERT models on both datasets.

Balancing data improves all transformer models, although the increase in performance is just 1-2\% (F1-score increase from 0.86 to 0.88 after five epochs on the T5 model, or 0.89-0.91 in DistilBERT). However, certain relationship types in the unbalanced dataset had a large gap between precision and recall (e.g. \textit{"No Explicit relationship"} in unbalanced had P=0.88, R=0.26), while in the balanced dataset precision and recall were closer (for the same class P=0.88, R=0.72).

\begin{table*}[!t]
    \centering
    \begin{tabular}{|l|l|l|l|l|l|l|}
    \hline
    	& \multicolumn{3}{|c|}{T5} & \multicolumn{3}{c|}{DistilBERT} \\
        Class & Precision & Recall & F1-score & Precision & Recall & F1-score \\ \hline
        \multicolumn{7}{|l|}{Unbalanced dataset}   \\  \hline
         Overall (weighted average) &   0.87 &  0.87  &  0.86 & 0.89 & 0.89 & 0.89 \\
        
        \hspace*{3mm} Biomarker & 1.00 &   0.52 &  0.69 & 0.75 & 0.63 & 0.69\\
        \hspace*{3mm} No Explicit Relationship   &  0.88  &  0.26  & 0.40 & 0.57 & 0.52 & 0.54 \\
        \hspace*{3mm} Plays a role &   0.91   &   0.95  &  0.93 & 0.96 & 0.95 & 0.95 \\
        \hspace*{3mm} Target-\textgreater Causative  & 0.84 & 0.90 &  0.87  & 0.89 & 0.94 & 0.91\\
        \hspace*{3mm} Target-\textgreater General        &  0.75 & 0.79 & 0.77 & 0.72 & 0.75 & 0.74\\
        \hspace*{3mm} Target-\textgreater Decrease Disease & 0.77 &   0.79 &  0.78  & 0.74 & 0.82 & 0.78\\
        \hspace*{3mm} Target-\textgreater Increase Disease   &  0.85   &   0.82   &   0.83 & 0.78 & 0.79 & 0.79\\
        \hline
        
         \multicolumn{7}{|l|}{Balanced dataset}  \\  \hline  
       Overall (weighted average)  &  0.88 & 0.88 & 0.88 & 0.91 & 0.91 & 0.91\\
        \hspace*{3mm} Biomarker & 0.97   &   0.95   &   0.96 & 0.91 & 0.93 & 0.92 \\
        \hspace*{3mm} No Explicit Relationship   & 0.88  &   0.72 &    0.79 & 0.92 & 0.86 & 0.89 \\
        \hspace*{3mm} Plays a role &  0.86   & 0.80 &   0.83 & 0.86 & .0.82 & 0.84\\ 
        \hspace*{3mm} Target-\textgreater Causative  & 0.90 &    0.96 &     0.93 & 0.97 & 0.95 & 0.96\\
        \hspace*{3mm} Target-\textgreater General        & 0.83   & 0.87 &    0.85 & 0.92 & 0.93 & 0.93\\
        \hspace*{3mm} Target-\textgreater  Decrease Disease & 0.83 &    0.91  & 0.87  & 0.84 & 0.93 & 0.88 \\
        \hspace*{3mm} Target-\textgreater  Increase Disease   &  0.91  &   0.95   &  0.93 & 0.91 & 0.91 & 0.91 \\ 
        \hline
        
    \end{tabular}
    \caption{Results of the best performing fine-tuned T5 and DistilBERT models (after 5 epochs)}
    \label{tab:t5_res}
\end{table*}

\begin{table*}[!t]
    \centering
    \begin{tabular}{|l|l|l|l|l|l|l|}
    \hline
    	& \multicolumn{3}{|c|}{SciFive} & \multicolumn{3}{c|}{PubMedBERT} \\
        Class & Precision & Recall & F1-score & Precision & Recall & F1-score \\ \hline
        \multicolumn{7}{|l|}{Unbalanced dataset}   \\  \hline
         Overall (weighted average) &   0.91 &  0.87  &  0.88 & 0.90 & 0.90 & 0.90 \\
        
        \hspace*{3mm} Biomarker & 0.87 &   0.62 &  0.72 & 0.76 & 0.90 & 0.83\\
        \hspace*{3mm} No Explicit Relationship   &  0.29  &  0.85  & 0.43 & 0.74 & 0.44 & 0.56 \\
        \hspace*{3mm} Plays a role &   0.97   &   0.93  &  0.94 & 0.95 & 0.96 & 0.95 \\
        \hspace*{3mm} Target-\textgreater Causative  & 0.87 & 0.93 &  0.90  & 0.87 & 0.84 & 0.85\\
        \hspace*{3mm} Target-\textgreater General        &  0.87 & 0.63 & 0.73 & 0.74 & 0.86 & 0.80\\
        \hspace*{3mm} Target-\textgreater Decrease Disease & 0.89 &   0.68 &  0.77  & 0.83 & 0.88 & 0.85\\
        \hspace*{3mm} Target-\textgreater Increase Disease   &  0.73   &   0.80   &   0.76 & 0.86 & 0.84 & 0.85\\
        \hline
        
         \multicolumn{7}{|l|}{Balanced dataset}  \\  \hline  
       Overall (weighted average)  &  0.90 & 0.88 & 0.89 & 0.92 & 0.92 & 0.92\\
        \hspace*{3mm} Biomarker & 1.00   &   0.91   &   0.95 & 0.97 & 0.97 & 0.97 \\
        \hspace*{3mm} No Explicit Relationship   & 0.70  &   0.86 &    0.77 & 0.95 & 0.88 & 0.92 \\
        \hspace*{3mm} Plays a role &  0.71   & 0.92 &   0.80 & 0.92 & 0.88 & 0.90\\ 
        \hspace*{3mm} Target-\textgreater Causative  & 0.99 &    0.89 &     0.94 & 0.94 & 0.96 & 0.95\\
        \hspace*{3mm} Target-\textgreater General        & 0.96   & 0.89 &    0.92 & 0.91 & 0.92 & 0.91\\
        \hspace*{3mm} Target-\textgreater  Decrease Disease & 0.91 &    0.80  & 0.85  & 0.88 & 0.92 & 0.90 \\
        \hspace*{3mm} Target-\textgreater  Increase Disease   &  0.96  &   0.92   &  0.94 & 0.91 & 0.93 & 0.92 \\ 
        \hline
        
    \end{tabular}
    \caption{Results of the best performing fine-tuned SciFive and PubMedBERT models (after 5 epochs)}
    \label{tab:t6_res}
\end{table*}  

We present the results of relationship classification after five epochs using general domain models in Table \ref{tab:t5_res}, while results for domain-specific models are in Table \ref{tab:t6_res}. Overall, the BERT-based models performed better on both data sets, even though the performance difference was just 2-3\%. Also, the BERT-based models performed better on the majority of relationship types. The stronger performance of DistilBERT, compared to T5-based models, is surprising and interesting due to its much smaller nature (66 million parameters in DistilBERT base compared to 220 million parameters in T5 base). This may be due to the multi-task and text-to-text nature of the T5 model, as a number of parameters need to be retained for other tasks and prefixes, as well as encoding to textual output. The best performing model was PubMedBERT, achieving F1-score of 0.92, followed by DistilBERT with F1-score of 0.91. The performance difference between PubMedBERT and DistilBERT is expected and within 3\% loss between BERT-based models and distilled version of it, that authors of original paper reported.  

We have added prefix into the T5 model and trained it for the classification of gene-associated modes of action into four possible classes: (1) activation, (2) inhibition, (3) mutation, and  (4) not reported. The utility of the T5 model is that a single model can perform both classifications of sentences by relationship type as well as the mode of action, for which we would need separate DistilBERT-based models.  The model was trained on the unbalanced and balanced dataset (each class containing 300 examples of each class). The model was trained for 5 epochs. The results are presented in Table \ref{tab:t5_res2}.

\begin{table*}[!t]
    \centering
    \begin{tabular}{|l|l|l|l|}
    \hline
        Class & Precision & Recall & F1-score \\ \hline
        \multicolumn{4}{|l|}{Unbalanced dataset}   \\  \hline
        \hspace*{3mm} Overall (weighted average) &   0.95 &  0.95  &  0.94  \\
        
        \hspace*{6mm} Activation & 1.00    &  1.00    &  1.00 \\
        \hspace*{6mm} Inhibition   & 0.87   &  1.00  &   0.93 \\
        \hspace*{6mm} Mutation &    1.00  &    0.77   &   0.87 \\
        \hspace*{6mm} Not reported  &  0.94   &   1.00   &   0.97\\

         \multicolumn{4}{|l|}{Balanced dataset (300 examples per class)}  \\  \hline  
                \hspace*{3mm} Overall (weighted average) &   0.98    & 0.97   &   0.97  \\
        
        \hspace*{6mm} Activation & 1.00    &  1.00    &  1.00 \\
        \hspace*{6mm} Inhibition   &    0.93   &  1.00   &  0.96  \\
        \hspace*{6mm} Mutation &    0.97   &  1.00   &   0.99 \\
        \hspace*{6mm} Not reported  &   1.00  &   0.90  &   0.95\\
        \hline
    \end{tabular}
    \caption{Results of the best performing mode-of-action T5 model}
    \label{tab:t5_res2}
\end{table*}  

Mode-of-action detection performs well with quite a small amount of data. This is because terms used for mode-of-action are in a relatively closed set (activation, inhibition, inhibitor, agonist, antagonist, mutation, modulation, etc.), and the language model is able to transfer and infer them from its pre-training on the C4 dataset. However, adding data helps improve it.

\section{Discussion}

The presented rule-based methodology is currently the base of the developed knowledge graph. With about 5 million typed relationships and over 30 million co-occurrences, it presents a powerful tool for drug discovery, target identification, indication expansion, and even pharmacovigilance. The graph structure allows for analysis over multiple hops. This will be further improved by adding protein-protein, drug-drug, and disease-disease interactions, on which we are working. It enables visualization of interaction pathways for diseases, graph learning for finding potentially missing relationships, and validating hypotheses about weak relationships. 

The current number of relationships in our graph is comparable with other state-of-the-art databases and graphs that consider the same or similar relationships. 	\cite{yang2021mining} developed a method for creating a stroke knowledge graph using PKDE4J based on 9 entity types and 30 relation types. The relationship extraction method based on BioBERT performed with an F1-score of 84.26\% and they managed to extract 157 000 relationships based on stroke-only papers in Pubmed (about 130 000 abstracts). \cite{lee2022bertsrc} developed likewise a method based on PKDE4J for entity identification and SciBERT for classification of relationships between genes, diseases, and compounds. The data model had 8 relationship types based on whether the relationship is directed, undirected, positive, negative, and has an increasing or decreasing effect on the object entity. The best performance of their model is an F1-score of 81.7\%. We believe that our model performed better (91\% F1-score), because the data model is more granular and crafted for particular entity pair relationships, therefore easier to learn than relationship types generic for any biomedical entity pair. 

\cite{kim2017analysis} focused on extracting gene-disease evidence sentences, using existing tools extracting genetic events, but did not classify the relationship between genes and diseases. They managed to extract about 7.3 million evidence sentences from PubMed. Our system is extracting mode of action, which partially compares to biological events, on top of which it also extracts typed relationships. Our system extracted 2.7 million typed relationships and 20 million co-occurrences, therefore both wider (co-occurrences) and more detailed (typed relationships with modes of action) evidence.

\cite{bhasuran2018automatic} used SVM on word embeddings to classify the existence of gene-disease relationships. The method was evaluated with an F1-score of about 83\%-87\%. RENET2 achieved 72.13\% F1-score for extracting gene-disease relationships from PubMedCentral articles \citep{su2021renet2}. However, gene-disease association types were not classified. Only the existence of the association was annotated. A number of methods were proposed for similar gene-disease association extraction without naming the relationship type based on DisGeNet dataset \citep{pinero2016disgenet,hebbar2021covidbert,parmar2020biomedical}. Since the publication of DisGeNet, the research in this area accelerated. Publicly available datasets with annotated biomedical relationship types are rare. Therefore, there was a need for the creation of new gene-disease relationship data set with our described data model. 

Our method based on PubMedBERT, as well as DistilBERT, has better results than most methods reported in the literature. However, the model relies on the relationship types, consistency of annotators, and size of the dataset. It is interesting to note that domain-specific models added a small increase in performance (1\% in both cases). While the difference between DistilBERT and PubMedBERT may be attributed to the knowledge distillation process, it is not the case for SciFive and T5 models, where increase certainly comes from domain specificity. However, since the meaning of a biomedical relationship is often described by terms and phrases also often used in the general domain, the effect of domain-specific models is not significant.


\section{Future work}

The creation of a comprehensive biomedical knowledge graph for target identification, indication expansion, and drug discovery is a long-term task. Some of the future activities on utilizing and improving our knowledge graph will involve: 
\begin{itemize}
\item \textbf{Develop machine learning, transformer-based models for other relationship types (drug-gene, drug-disease, drug-drug, gene-gene, disease-disease)}. This may involve further annotation of data for other relationship pairs and creating a model based on these annotations. 
\item \textbf{Unifying relationships obtained from unstructured (literature, clinical trials, expert reports, grant proposals) and structured data sources} - Combining structured and unstructured data provides better quality of results and opens the possibility for a more detailed and comprehensive evaluation of links in the graph. 
\item \textbf{Developing an interface for exploring relationships and their evidence} - graphical user interface would enable a wider scientific audience to utilize the graph. This is especially important due to the fact that the majority of pharmacologists and biologists working in the pharmaceutical industry do not have an extensive computational background. This would allow them to be more efficient in generating and evaluating hypotheses before going to the laboratory. 
\item \textbf{Predicting novel target candidates using graph and temporal graph learning methods} - based on the chronology of the appearance of relationships in the graph, it may be possible to learn patterns and predict relationships between entities that would be identified in future. Based on the year of publication, we can track when certain relationships appeared in literature and how evidence around it mounted. Therefore, it would be possible to automatically generate a hypothesis about the existence of yet undiscovered biological relationships using temporal graph neural networks \citep{wang2020traffic}.
\end{itemize}

\section{Conclusions}

In this paper, we have presented one rule-based approach, that mainly served as a starting point for obtaining biomedical relationship data. Further, we have compared traditional machine learning approaches, with modern, state-of-the-art language models and transformer approaches (DistilBERT, PubMedBERT, T5 and SciFive models). 

In all approaches, the improvement was noticeable with balanced datasets, however, fine-tuned transformer-based models (DistilBERT, PubMedBERT, T5, and SciFive) were more resilient and did not depend so much on balanced data sets as some older and traditional approaches would (Naive Bayer and Random Forests). Also, transformer-based models, due to their pre-training on larger data, are able to generalize well from a fairly small amount of data. BERT-based models (PubMedBERT and DistilBERT) performed slightly better than the T5 models (T5 base and SciFive base), which was a surprising finding since T5  has about 4 times more parameters than DistilBERT and about 2 times more than PubMedBERT. However, this may be due to the multi-task nature of T5, as well as the fact that part of the model has to be used for text-to-text generation.

Developing machine learning data sets for tasks such as relationship extraction can be quite expensive. On the market, the pricing of a single annotated sentence can range between 1-3 euros, depending on the complexity of the task. However, this quickly scales, once the data set has 7 annotation classes and there is a need for over 1000 examples per annotation class in the data set. The commissioned manual annotations of our data set (around 7,000 sentences in total) cost 16,200 euros. The further cost comes from cloud infrastructure and machine learning engineering. Costs in developing relationship extraction models and approaches remain one of the main challenges.

Nevertheless, fine-tuning transformer models proved to be a promising approach. First of all, the performance of the model outperformed all other approaches, with over 92\% F1-score overall in the case of PubMedBERT, and with the majority of annotation classes breaching 85\% F1-score. A review of the literature showed that the model performance is state-of-the-art for biomedical typed relationship extraction. Also, the model showed stability in terms of both precision and recall (unlike the rule-based approach, which has high precision but fairly low recall). On the other hand, T5 models are multi-task models, where it is possible to successfully address multiple problems with a single model, which makes valuable savings in terms of managing and updating the models. Lastly, fine-tuned T5 models, as they are text-to-text models, are easy to use and data preparation is kept simple. Our evaluation also showed, that in the particular case of gene-disease relationship extraction, domain-specific models add little performance boost.

In terms of limitation, T5 models are large, multi-task models, whose base model contains 220 million parameters. This is, for example, twice the size of the PubMedBERT base and over four times the size of the DistilBERT model, and it contributes heavily to the speed of fine-tuning and execution, making the processing slow. Despite the fact that DistilBERT can be trained only on a single task and there is a need for post-processing of outputs, the model has both performance and speed advantages compared to the T5-based models. DistilBERT was outperformed by PubMedBERT by 1\%. However, due to DistilBERTs size and speed advantages, it may be preferable for many productional use-cases.




\bibliographystyle{elsarticle-harv} 
\bibliography{knowledge_graph_paper2}
\end{document}